# Simple2Complex: Global Optimization by Gradient Descent

Ming Li
lmcsu@sina.com

*Abstract*—A method named simple2complex for modeling and training deep neural networks is proposed. Simple2complex train deep neural networks by smoothly adding more and more layers to the shallow networks, as the learning procedure going on, the network is just like growing. Compared with learning by end2end, simple2complex is with less possibility trapping into local minimal, namely, owning ability for global optimization. Cifar10 is used for verifying the superiority of simple2complex.

*Keywords: Simple to Complex; Global Gradient Optimization; Series Neural Network; Deep Learning; Artificial Intelligence;*

## I. INTRODUCTION

Deep learning has dramatically advanced the state of the art in many fields such as vision, NLP, games and so on. Stochastic gradient descent (SGD) and it's variants such as Momentum and Adagrad used to achieve state of the art performance for many problems have been proved a effective way of training deep neural networks[1, 2]. In recent years, it is popular to model a complex problem as a extremely deep neural network and train it by SGD in the end to end way. It is well known that SGD is an optimization method based on gradient descent, and optimization by gradient decent end to end is prone to fall into local minimum as the nonlinearity of parameter space become more and more higher. Though alleviated by batch normalization[3] and residual network[4, 5, 6], it still become more and more difficult as the depth of neural network increasing. Therefore, the questions arises: is there exists a global optimization method based on gradient descent and is there a better choice than training by end to end ?

In this paper, the above questions are answered by proposing a method named simple2complex for modeling and training deep neural networks.

## II. PHILOSOPHICAL AND BIOLOGICAL VIEW

The world is complex and highly non-convex, but if we look over it by a hierarchical way "from global to local" or "from abstract to detail", a simple and relative convex world is presented to you. "Do not know the true face of Mount Lu just because in the mountains". Imaging we are birds flying high over the mountain and look down at it, then we can get it's global appearance, and we can easily find out the approximate region that the lowest valley located at. Then, we fly to the region and slightly decline flight attitude to see the region more clearly, therefore, we can further reduce the region scope. This process move in circles until we find the exact position of that valley, and the road to the Arcadia is right over there.

Considering the long biological evolution history as a whole optimization process is very interesting. The process begun with the occurrence of multi-molecular system about 40 hundred-million years ago, then be prokaryote, eucaryon, invertebrate, fish, reptile, bird, mammal, ape and the last is human. Early species possesses simple stimuli-responsive molecular structure as original prototype of neural system, under the rigorous long time natural selection, the original prototype evolved into shallow neural system, at this stage, the fundamental global biological structure is formed. Then, induced by various kinds of stimulus such as light, voice, chemical ingredients , neurons close to those stimulus continuously divide and grow, as neurons grow toward different stimulus, kinds of sub-neuron-structures is formed, finally those sub-neuron-structures evolved into eyes, ears, nose, etc. As the neural systems grow deeper and deeper, advanced species arose with the assistance of nature selection. This process continues for hundreds of millions of years, finally human beings appeared. This is a typical simple2complex optimization procedure, first, learn a shallow neural systems to model basic global biological functions, then, let the neural systems grow deeper and deeper to model more and more complicated biological functions. On the contrary, if we image biological evolution history as a end2end optimization procedure, then, at the beginning, there should be a huge deep neural networks initialized by lots of random inorganics, optimizing this systems by nature must be much harder than the way of simple2complex described earlier.

## III. MATHMAMTIC VIEW

We always model a complex problem by a highly non-convex functions $f_c(x)$ with a lot of parameters, then search over those parameters using gradient, and obviously, always trapped into local minimal. In this paper, we model a complex problem by a sequence of functions $f_{s_0}(x)$, $f_{s_1}(x)$, $f_{s_2}(x)$, ..., $f_{s_n}(x)$, in which, $f_{s_0}$ is a relative simple and convex function with few parameters, and $f_{s_{i+1}}$ is a little more complicated than $f_{s_i}$ with a little more parameters, not strictly, $f_{s_{i+1}}(x)$ can be seen as $f_{s_i}(x)$ soft-add $f_{r_i}(x)$, $f_{r_i}$ is a residual function between $f_{s_i}$ and $f_{s_{i+1}}$. And $f_{s_n}$ is the same function compared with $f_c(x)$ which is more and more capable of learning the ground truth as n incremented. Parameters of $f_c$, $f_{s_i}$ and $f_{r_i}$ are denoted as

$w_c$, $w_{s_i}$ and $w_{r_i}$ respectively, and comply with the following conditions:

$$w_{s_{i+1}} = w_{s_i} \cup w_{r_i} \quad (1)$$

$$w_c = w_{s_n} \quad (2)$$

The difference between end2end and simple2complex is that end2end optimize $f_c(x)$ over $w_c$ directly, but simple2complex firstly optimize $f_{s_0}(x)$ over $w_{s_0}$ to learn the coarse global outline of ground truth, after $f_{s_i}(x)$ has been optimized over $w_{s_i}$ to learn a relative coarse outline, $f_{s_i}(x)$ is used to direct the optimization of $f_{r_i}(x)$ over $w_{r_i}$, and at the same time, the learned residual function $f_{r_i}(x)$ then can be used to refine $f_{s_i}(x)$, in this way, $f_{s_{i+1}}(x)$ is been optimized over $w_{s_{i+1}} = w_{s_i} \cup w_{r_i}$ to learn a relative fine outline.

$f_c$ is a function of $x$, and also, it's a function of $w_c$. For the case that $f_c$ is represented by a deep neural network, $f_c(x)$ and $f_c(w_c)$ have similar shape and properties because of the symmetry of $x$ and $w_c$ occurring in $f_c$. Optimization of $f_c(x)$ over $w_c$ is equal to conducting gradient descent upon $f_c(w_c)$ which owning similar shape and complexity of $f_c(x)$. And the argument above is the same for $f_{s_i}$ and $f_{r_i}$. $f_c(x)$ is a highly non-convex functions, therefore optimizing $f_c(x)$ over $w_c$ by end2end is a highly non-convex optimization problem, optimization will be harder and harder because $f_c(x)$ and $f_c(w_c)$ become more and more complex as the neural network become deeper and deeper. But for simple2complex, the problem above is not that serious as end2end. $f_{s_0}(x)$ is a relative simple and convex function, so optimization of $f_{s_0}(x)$ over $w_{s_0}$ is much easier than optimization of $f_c(x)$, so it's more likely to achieve global optimum. Because the global optimum of $f_{s_0}(x)$ means coarse global outline of ground truth, so this global optimum of $f_{s_0}(x)$ is near to the global optimum of $f_{s_n}(x)$. When optimizing $f_{s_{i+1}}(x)$, we just need to learn a simple residual function $f_{r_i}(x)$ to refine the already learned $f_{s_i}(x)$, this is much easier than learning $f_{s_{i+1}}(x)$ by end2end, because the global optimum of $f_{s_i}(x)$ is close to global optimum of $f_{s_{i+1}}(x)$, so if optimization of $f_{s_i}(x)$ achieve global optimum, then optimization of $f_{s_{i+1}}(x)$ can easily achieve global optimum by slightly adjust the global optimum parameters of $f_{s_i}(x)$. By this way, simple2complex is expected to achieve global optimum of $f_{s_n}(x)$ which is also the global optimum of $f_c(x)$ with much greater chance than end2end.

## IV. METHODS

For simplicity, the method of simple to complex is denoted as s2c, and the method of end to end is denoted as e2e. $f_{s_0}(x)$ is represented by a plain neural network $N_{s_0}$ which is constructed by $t$ layers where $t$ is a small number. The network $N_{s_{i+1}}$ representing $f_{s_{i+1}}(x)$ is constructed by adding 2 new layers as residual path for each new layer added to $N_{s_i}$. Optimization of $N_{s_1}, N_{s_2}, \ldots, N_{s_n}$ is conducted sequentially after $N_{s_0}$ by SGD with fixed learning rate. After $N_{s_i}$ is been optimized, optimization of $N_{s_{i+1}}$ is done as following: firstly, parameters of layers of $N_{s_{i+1}}$ which is inherited from $N_{s_i}$ is restored from optimized $N_{s_i}$, and the parameters of the new added layers is carefully initialized that keep the initial function of $N_{s_{i+1}}$ equals to the optimized $N_{s_i}$, secondly, standard SGD is performed on $N_{s_{i+1}}$ with the fixed learning rate until convergence. After $N_{s_n}$ is optimized as above, then SGD will still be performed on $N_{s_n}$ for more iterations by reducing learning rate gradually, and then, the final optimized $N_{s_n}$ is what we want.

As shown in Figure 1, a directed edge is used to represent a conv-bn layer, one or several directed edge emitting from one point means data split, one or several directed edge directing to one point means element-wise add operation followed by a nonlinear activation layer. Left graph in Figure 1 represents $N_{s_0}$ with only one layer, right graph in Figure 1 represents $N_{s_n}$, this type of neural network can be called series neural network because that any neuron in the network can be seen as mathematical series, the type of series depends on the type of nonlinear activation layer, it can be called Fourier series neural network if trigonometric function such as tanh is used and Taylor series neural network if power function such as $y = x^2$ is used. In this paper, relu series neural network is used in all experiments. The value of gamma of batch normalization can be seen as coefficient of corresponding component in the series. Decomposing a function into a series may be a more appropriate and general way than residual method.

To guarantee the initial function of $N_{s_{i+1}}$ equals to the optimized $N_{s_i}$, the gamma of the second batch normalization layer of the new-added two-layer residual paths is initialized as zero, the initialization method for other parameters of s2c is the same as e2e.

Standard SGD with momentum 0.9, weight decay 0.0002 and batch size 128 is used for s2c and e2e, 0.1 is used as the fixed learning rate for s2c and initial learning rate for e2e. Learning rate is multiplied by 0.5 once loss stagnate when train by e2e and train final $N_{s_n}$ by s2c ( 60000 steps for growing $N_{s_i}$ to $N_{s_{i+1}}$, 80000 steps for learning rate 0.1 when train by e2e and train final $N_{s_n}$ by s2c, 40000 steps for decaying learning rate ). Moving average 0.9999 is used for bn and all trainable variables for boosting performance.

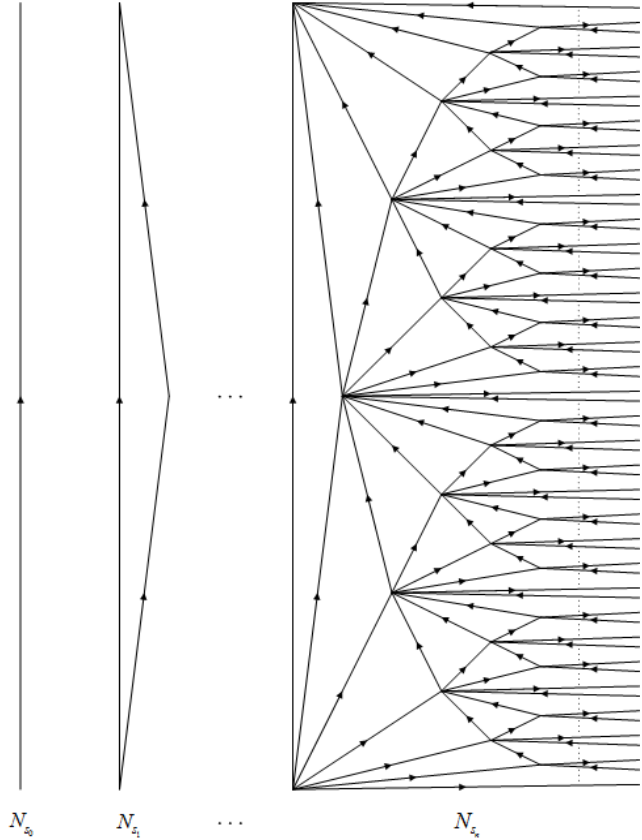

Figure 1. $N_{s_0}$ with only one layer growing to a series neural network $N_{s_n}$

V. EXPERIMENTS

For saving experiment time, a very small but still deep series neural network with only 8, 16 or 32 filters per layer is designed for cifar10 as shown in figure 2. The left graph is a six layer plain network, namely, $N_{s_0}$. The right graph is $N_{s_n}$ where $n$ equals to 2. The kernel size and padding of each convolution layer is elaborately chosen to guarantee each input element of add operation layer having the same receptive field. Starting from only one layer may be more ideal, but complicated tricks such as pooling before conv is needed for avoiding big kernel, so just left it for future work.

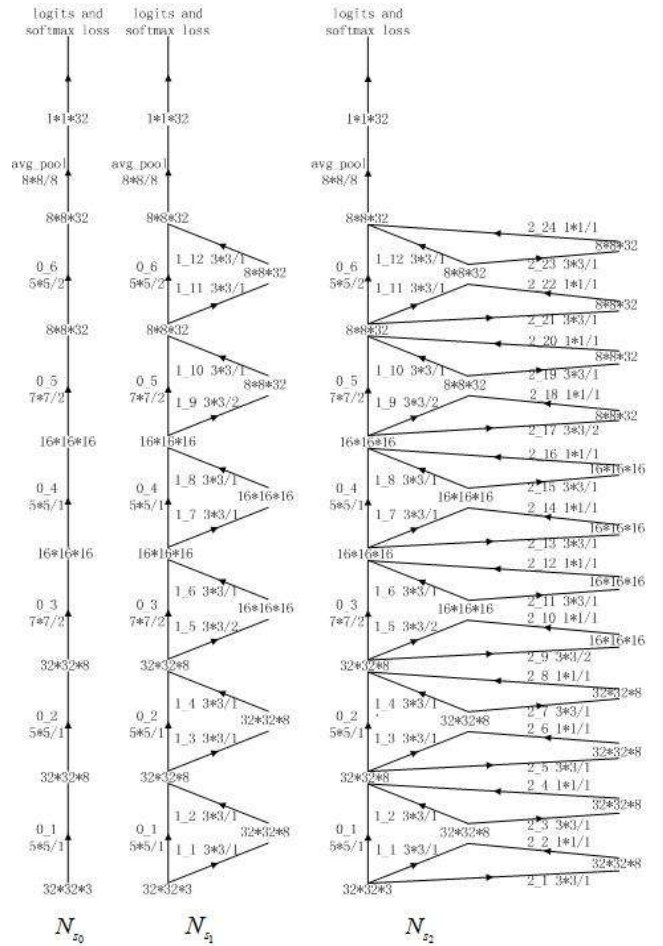

Figure 2. neural network designed for cifar10

The cifar10 testing and training accuracy at the final iteration step of each learning rate by e2e and s2c for $N_{s_n}$ is shown in Table 1. It is not fair to compare e2e and s2c because that structure of series neural network may make behavior of training by e2e imply the thoughts of simple2complex, while, the accuracy of s2c is still higher than e2e obviously both for testing and training accuracy. Besides, the difference between testing and training accuracy of s2c is smaller than e2e which means s2c is not easy to over-fitting as e2e.

The cifar10 training and testing accuracy of optimized $N_{s_0}$, $N_{s_1}$, $N_{s_2}$ by s2c on the fixed learning rate 0.1 is shown in Table 2. With the growth of the network, accuracy become more and more high. Therefore, probably, series network with arbitrary depth can be trained by s2c without obstacles.

Table 1. cifar10 accuracy of e2e and s2c

| lr | e2e | | | s2c | | |
|---|---|---|---|---|---|---|
| | train | test | trn-tst | train | test | trn-tst |
| 0.1 | 0.9370 | 0.8925 | 0.0445 | 0.9434 | 0.9000 | 0.0434 |
| 0.05 | 0.9604 | 0.9028 | 0.0576 | 0.9634 | 0.9090 | 0.0544 |
| 0.025 | 0.9767 | **0.9057** | 0.0710 | 0.9801 | 0.9093 | 0.0708 |
| 0.0125 | 0.9849 | 0.9032 | 0.0817 | 0.9897 | **0.9105** | 0.0792 |
| 0.00625 | 0.9916 | 0.9023 | 0.0893 | 0.9961 | 0.9075 | 0.0886 |

Table 2. cifar10 accuracy during growing

| | train | test |
|---|---|---|
| $N_{s_0}$ | 0.8793 | 0.8372 |
| $N_{s_1}$ | 0.9195 | 0.8750 |
| $N_{s_2}$ | 0.9434 | 0.9000 |

A nonsingular function can be decomposed as a series such as Fourier series or Taylor series, and the absolute value of coefficient of each components in the series decreases as the frequency of component increases commonly. Therefore, the gamma values under some add operations at highest testing accuracy iteration step is reported to verify the above property of series. Table 3, Table 4, Table 5 show $avg_{chanel}(|gamma|)$ of some add operations for s2c and e2e. Table 6 show all gamma values of all channels of layer 0_6, 1_12, 2_24 for s2c. As shown in these Tables, Gamma values are accord with the above property of series either by s2c or e2e, but gamma values of e2e decease not as obviously as s2c which means s2c is more able to learn the series decomposition.

Table 3. $avg_{chanel}(|gamma|)$ of 0_6, 1_12 and 2_24

| s2c | | | e2e | | |
|---|---|---|---|---|---|
| 0_6 | 1_12 | 2_24 | 0_6 | 1_12 | 2_24 |
| 9.795 | 4.040 | 2.580 | 6.172 | 3.028 | 2.503 |

Table 4. $avg_{chanel}(|gamma|)$ of 0_3, 1_6 and 2_12

| s2c | | | e2e | | |
|---|---|---|---|---|---|
| 0_3 | 1_6 | 2_12 | 0_3 | 1_6 | 2_12 |
| 4.416 | 2.130 | 1.653 | 3.370 | 1.538 | 0.991 |

Table 5. $avg_{chanel}(|gamma|)$ of 1_5 and 2_10

| s2c | | e2e | |
|---|---|---|---|
| 1_5 | 2_10 | 1_5 | 2_10 |
| 2.851 | 0.859 | 2.247 | 0.946 |

The absolute gamma values of new added layers decreasing to zero can be considered as a criteria of stopping growing because growing more layers will not bring any performance gain any more. Analyzing the gamma values of Table 3-Table 6, the network is far from achieving performance bottleneck, and growing more layers continuously will still obtain accuracy gain anticipatively but it is not done in this paper.

Table 6. all gamma values of layer 0_6, 1_12, 2_24

| channel index | 0_6 | 1_12 | 2_24 |
|---|---|---|---|
| 0 | 11.878 | 4.059 | 2.098 |
| 1 | 11.224 | 4.896 | -4.183 |
| 2 | 10.714 | -4.048 | -2.876 |
| 3 | 10.320 | 4.186 | -0.323 |
| 4 | 10.075 | -4.154 | -3.853 |
| 5 | 11.035 | 4.414 | -1.865 |
| 6 | 9.911 | -3.128 | -3.831 |
| 7 | 8.423 | 5.592 | 3.319 |
| 8 | 10.398 | 4.419 | -0.796 |
| 9 | 9.296 | 2.820 | 2.611 |
| 10 | 10.923 | -2.663 | -3.371 |
| 11 | 7.598 | 2.721 | -2.791 |
| 12 | 8.586 | -4.664 | -1.894 |
| 13 | 10.684 | 4.030 | -3.511 |
| 14 | 8.916 | 3.758 | -1.303 |
| 15 | 9.846 | -3.805 | -3.513 |
| 16 | 8.652 | 2.754 | -1.427 |
| 17 | 10.647 | -5.310 | 2.886 |
| 18 | 11.171 | 4.758 | -4.434 |
| 19 | 10.769 | 4.304 | -1.393 |
| 20 | 9.545 | 5.662 | 2.969 |
| 21 | 8.743 | -3.815 | -2.569 |
| 22 | 8.641 | 3.263 | 2.378 |
| 23 | 10.744 | -4.985 | 2.204 |
| 24 | 7.952 | -3.736 | -1.717 |
| 25 | 10.103 | -4.709 | -3.056 |
| 26 | 9.117 | 3.626 | -1.258 |
| 27 | 8.609 | -3.517 | 1.550 |
| 28 | 11.358 | 3.784 | 3.750 |
| 29 | 10.055 | -4.935 | -2.931 |
| 30 | 8.810 | 3.118 | 3.406 |
| 31 | 8.694 | -3.620 | -2.492 |

## VI. CONCLUSIONS

A method called simple2complex is proposed aiming to use gradient decent for global optimization. The superiority of simple2complex is verified by experiments on cifar10.